\documentclass{article}

\PassOptionsToPackage{numbers, compress}{natbib}

\usepackage{PRIMEarxiv}
\usepackage{color}
\usepackage{natbib}
\usepackage{wrapfig}

\usepackage{amssymb}
\usepackage[utf8]{inputenc}
\usepackage[T1]{fontenc}
\usepackage{hyperref}
\usepackage{url}
\usepackage{booktabs}
\usepackage{amsfonts}
\usepackage{nicefrac}
\usepackage{microtype}
\usepackage{array}
\usepackage{multicol}
\usepackage{multirow}
\usepackage[utf8]{inputenc}
\usepackage{tabularx}
\usepackage[T1]{fontenc}
\usepackage{graphicx}
\usepackage{enumitem}
\usepackage[table,xcdraw]{xcolor}
\usepackage{amsmath}
\usepackage{fontawesome}
\usepackage{makecell}

\usepackage[utf8]{inputenc}

\definecolor{bb}{rgb}{0.12,0.565,1}
\definecolor{gg}{rgb}{0.2,0.8,0.2}
\definecolor{rr}{rgb}{1,0.85,0.2}

\newif\ifdraft
\drafttrue

\hypersetup{
    colorlinks=true,
}

\newcommand{\ours}[0]{Baichuan-Audio\ }
\title{Baichuan-Audio: A Unified Framework for \\[8pt] End-to-End Speech Interaction}

\author{
    Tianpeng Li\enskip\enskip
    Jun Liu\enskip\enskip
    Tao Zhang\enskip\enskip
    Yuanbo Fang\enskip\enskip
    Da Pan\enskip\enskip 
    Mingrui Wang\enskip\enskip \\
    Zheng Liang\enskip\enskip
    Zehuan Li\enskip\enskip
    Mingan Lin\enskip\enskip
    Guosheng Dong\enskip\enskip 
    Jianhua Xu\enskip\enskip \\
    Haoze Sun\thanks{Corresponding author.}\enskip\enskip
    Zenan Zhou$\footnotemark[1]$\enskip\enskip
    Weipeng Chen\enskip\enskip 
    \vspace{0.2cm}
    \\
    Baichuan Inc. \enskip
    \\
    \texttt{\{sunhaoze, zhouzenan\}@baichuan-inc.com}
    \enskip \\
}

\begin{document}

\maketitle

\begin{abstract}
We introduce Baichuan-Audio, an end-to-end audio large language model that seamlessly integrates audio understanding and generation. It features a text-guided aligned speech generation mechanism, enabling real-time speech interaction with both comprehension and generation capabilities. Baichuan-Audio leverages a pre-trained ASR model, followed by multi-codebook discretization of speech at a frame rate of 12.5 Hz. This multi-codebook setup ensures that speech tokens retain both semantic and acoustic information. To further enhance modeling, an independent audio head is employed to process audio tokens, effectively capturing their unique characteristics.
To mitigate the loss of intelligence during pre-training and preserve the original capabilities of the LLM, we propose a two-stage pre-training strategy that maintains language understanding while enhancing audio modeling. Following alignment, the model excels in real-time speech-based conversation and exhibits outstanding question-answering capabilities, demonstrating its versatility and efficiency.
The proposed model demonstrates superior performance in real-time spoken dialogue and exhibits strong question-answering abilities. Our code, model and training data are available at \url{https://github.com/baichuan-inc/Baichuan-Audio}
\end{abstract}
\section{Introduction}

The development of audio dialogue models has shifted from traditional cascade modeling approaches to end-to-end audio large models \cite{hassid2024textually}. Traditional audio dialogue models operate through a sequential framework, where audio input is processed via Automatic Speech Recognition (ASR) to generate text. This text is then used by a language model to produce responses, which are converted into synthesized speech through Text-to-Speech (TTS) systems. While this approach leverages the capabilities of large language models (LLMs), it overlooks the influence of paralinguistic information and introduces processing delays along with errors accumulated through multiple data conversions. This step-by-step processing chain exacerbates these issues, impacting overall performance. In contrast, contemporary end-to-end audio large models, such as GPT-4o \cite{HelloGPT4o}, adopt a unified framework to directly process audio input, streamlining the task flow and enhancing both understanding and generation of audio content. This evolution facilitates more natural, fluid audio interactions and improves real-time performance.

Recently, open-source end-to-end audio interaction systems have advanced toward achieving real-time performance. These systems can be categorized into three main types. The first, exemplified by Freeze-Omni \cite{wang2024freeze}, aligns audio and text modalities through modality adapters, enabling pre-trained large language models to process audio inputs, with hidden states converted into speech waveforms via a speech decoder. While this approach preserves the original capabilities of LLMs, it lacks a holistic end-to-end understanding of the audio modality. The second type, represented by Moshi \cite{defossez2024moshi}, utilizes discrete audio tokens as input and adopts a multi-stream architecture to output both audio and text concurrently. The third type, such as GLM-4-Voice \cite{zeng2024glm}, differs by generating interleaved audio and text outputs, allowing text to guide audio generation for enhanced output quality. However, a key challenge for the second and third types lies in the integration of audio modalities, which often results in a noticeable reduction in reasoning capabilities compared to textual large language models.

In this work, we introduce Baichuan-Audio, an end-to-end audio large language model designed for real-time speech interaction. Similar to Moshi and GLM-4-Voice, Baichuan-Audio extends pre-trained LLMs to enable end-to-end audio input and output. This is achieved through the integration of the Baichuan-Audio-Tokenizer and a stream-matching decoder, which discretize audio signals into tokens and decode audio tokens back into speech waveforms, respectively. The tokenizer operates at a frame rate of 12.5 Hz and employs multi-codebook discretization to retain both semantic and acoustic information, enabling effective modeling of the speech modality within the LLM. Baichuan-Audio further incorporates an independent audio head to enhance the capability of model to process and capture unique audio features. We conducted large-scale pretraining on audio-text data comprising approximately 100 billion tokens. Based on an extensive audio corpus comprising 887k hours, we implement an interleaved data processing approach, drawing upon the methodology outlined in \cite{kim2024unified,nguyen2025spirit,zeng2024scaling}, to facilitate effective knowledge transfer within the LLM framework. To preserve textual understanding during audio modeling, a two-stage pretraining strategy was introduced, where audio embeddings and the audio head were initially trained independently to maintain language comprehension. Baichuan-Audio demonstrates exceptional performance in real-time speech interactions and exhibits robust question-answering capabilities, highlighting its versatility and efficiency.
The main contributions of \textbf{Baichuan-Audio} can be summarized as follows: 
\begin{itemize}

    \item \textbf{Unified and Outstanding Speech Capabilities}: We design an 8-layer RVQ audio tokenizer (Baichuan-Audio-Tokenizer) achieves an optimal balance between capturing semantic and acoustic information with 12.5 Hz frame rate, which supports high-quality controllable bilingual (Chinese and English) real-time conversations.

    \item \textbf{End-to-end Speech Interaction}: Baichuan-Audio is designed to process text, audio inputs, delivering high-quality text and speech outputs. It is capable of delivering high-quality, seamless speech interaction while maintaining intelligent response. At the same time, we have also open-sourced the training data and foundational model, providing valuable resources and tools to advance research and innovation in the field of voice interaction.
\end{itemize}

\section{Related works}
\subsection{Audio Large Language Models}
The development of audio large language models is driven by advancements in speech tokenizers and large language model research. The speech tokenizers serve as a crucial bridge between audio segments and discrete language models by transforming continuous audio signals into discrete tokens. The self-supervised learning models, such as HuBERT\cite{hsu2021hubert} and WavLM\cite{chen2022wavlm}, effectively capture semantic information from speech. The neural acoustic codecs \cite{zeghidour2021soundstream,defossez2022high,kumar2024high} are designed to preserve the full range of audio signal information through discrete encoding. Recent studies \cite{zhang2024speechtokenizer,defossez2024moshi} have also utilized distilling semantic features sush as pre-trained HuBERT to ensure that specific layers of residual vector quantization (RVQ) retain enriched semantic content. In audio understanding, methods like Qwen-Audio\cite{chu2023qwen,chu2024qwen2}, Wavllm\cite{hu2024wavllm}, and Pengi\cite{deshmukh2023pengi} combine Whisper encoder\cite{radford2023robust} with large language models, utilizing multi-task learning strategies across speech and language tasks. These approaches have demonstrated excellent performance in speech-to-text tasks. In text-to-speech (TTS), Wang et al. introduced VALL-E\cite{wang2023neural}, conceptualizing TTS as a conditional language modeling problem. To further enhance the efficiency and fidelity of speech synthesis, decoders such as CosyVoice\cite{du2024cosyvoice} and Matcha-TTS\cite{mehta2024matcha} leverage flow-matching techniques, enabling high-quality speech generation.

\subsection{End-to-end audio LLM}
End-to-end speech interaction models have emerged as a central research focus within the speech processing community, driven by the increasing demand for seamless and efficient multimodal communication systems. Inspired by GPT-4o\cite{HelloGPT4o}, significant advancements have been made in the development of open-source models. For instance, Moshi \cite{defossez2024moshi} introduces an end-to-end full-duplex spoken dialogue foundation model, capable of simultaneously generating audio tokens and text tokens through a multi-stream output mechanism. GLM-4-Voice \cite{zeng2024glm} leverages interleaved data for pre-training to enable text-guided interleaved generation of speech. Freeze-Omni \cite{wang2024freeze} extends the capabilities of large language models (LLMs) to process speech modalities by incorporating modality-specific adapters. It further employs a speech decoder to transform the LLM's output into audio tokens, enabling high-quality speech generation. Following a similar technical approach to Freeze-Omni, models such as VITA-1.5 \cite{fu2025vita} and SALMONN-Omni \cite{yu2024salmonn}. These developments collectively highlight the ongoing progress in creating robust and versatile frameworks for speech-based applications.

\section{\ours}
\begin{figure*}[t]
  \centering
  \includegraphics[width=16cm]{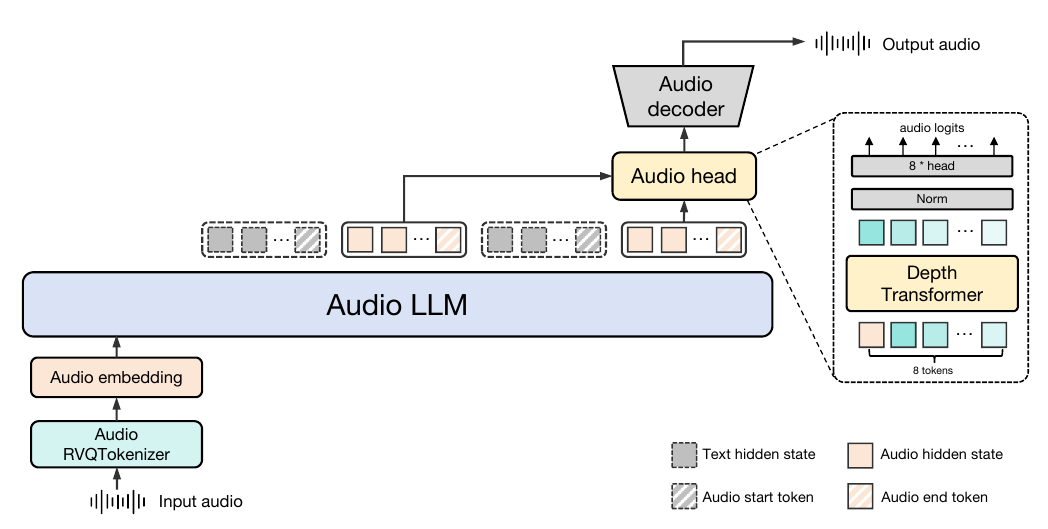}
  \caption{\textbf{The overview of Baichuan-Audio.} Our model is an end-to-end large audio  language model. When generating audio, the audio LLM alternately predicts text tokens and audio tokens. The audio tokens are then decoded by the flow-matching based audio decoder to produce the final audio.}
  \label{fig:audio}
\end{figure*}

Baichuan-Audio is an advanced end-to-end audio large language model designed for real-time speech interaction. The overall architecture of the Baichuan-Audio model is presented in \autoref{fig:audio}. The model architecture is structured around three foundational components: the Baichuan-Audio Tokenizer, the audio LLM and an audio decoder. The processing pipeline begins with the audio tokenizer, which converts raw audio input into discrete tokens by capturing both semantic and acoustic information. This is achieved through a combination of high-level feature extraction via the Whisper Encoder and RVQ techniques. The audio LLM then generates aligned text and audio tokens in an alternating manner, facilitated by a specialized token that enables seamless modality switching between text and audio. The audio tokens are subsequently processed by an independent audio head. Finally, a flow-matching based audio decoder reconstructs a high-quality Mel spectrogram from these tokens, which is then converted into an audio waveform via a vocoder. In this section, we will delve into the model architecture of Baichuan-Audio, providing a comprehensive analysis of its training data and the methodologies employed in its development.

\subsection{Audio Tokenization}

\par
\begin{wrapfigure}{r}{0.45\textwidth}
  \centering
  \includegraphics[width=6cm]{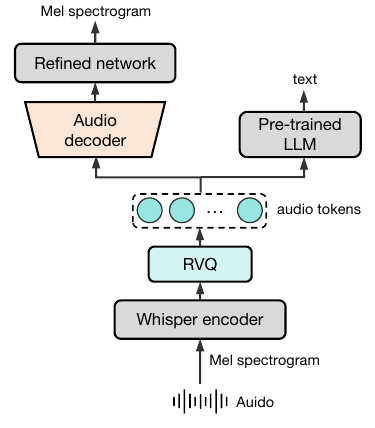}
  \caption{Baichuan-Audio-Tokenizer.}
  \label{fig:tokenizer}
\end{wrapfigure}

 The main challenge for current audio tokenizers lies in achieving an optimal balance between capturing the semantic and acoustic information in the input speech signal \cite{wang2024speech}. We believe that models such as Baichuan-Omni and Qwen-Audio provide a more direct approach to capturing semantic features, compared to self-supervised learning methods like HuBERT \cite{hsu2021hubert}. Meanwhile, audio tokenizer like Encodec \cite{defossez2022high} and SpeechTokenizer \cite{zhang2024speechtokenizer} are particularly effective at fully reconstructing audio features. To combine the advantages of these two approaches and inspired by the work of \cite{wu2024vila}, we propose Baichuan-Audio-Tokenizer, an audio tokenizer based on RVQ \cite{defossez2022high} and multi-objective training, as illustrated in \autoref{fig:tokenizer}. The Baichuan-Audio-Tokenizer retains the audio encoder and the LLM component from Baichuan-Omini \cite{li2024baichuan}, while adding an audio decoder structure after the encoder to reconstruct the input Mel spectrogram. The audio tokenizer is trained using a multi-objective optimization approach to effectively capture both semantic and acoustic information.

The Baichuan-Audio-Tokenizer utilizes a frame rate design of 12.5 tokens per second. The high-level audio features are extracted from Mel spectrograms using the Whisper Large Encoder~\cite{radford2022robustspeechrecognitionlargescale}, followed by a residual convolutional network performing 4$\times$ downsampling to obtain low frame rate audio features. Since the audio features output by the Whisper Encoder are high-dimensional, it is essential to use an 8-layer RVQ to minimize information loss during the quantization process. We design the codebook sizes in a decreasing manner as $\{8K, 4K, 2K, 1K, 1K, 1K, 1K, 1K\}$. The audio decoder employs a fully symmetric structure to the Whisper Encoder, processing its input through a deconvolution module for 4× upsampling. After passing through a series of Transformer layers, the sequence undergoes an additional 2× upsampling, resulting in a coarse Mel-spectrogram representation with 100 tokens per second. Inspired by \cite{li2019neural,meng2024autoregressive}, we design a refined network to improve the accuracy of Mel-spectrogram reconstruction, yielding high-quality refined Mel-spectrogram features. In the design of the audio reconstruction loss, we refer to \cite{meng2024autoregressive} and adopt a combination of L2 and L1 losses as the reconstruction loss. The reconstruction loss is defined as follows:
\begin{equation}
\begin{aligned}
  \text{Loss}_{reconstruct} &= L_1(\text{Mel}_{gt}, \text{Mel}_{coarse}) + L_1(\text{Mel}_{gt}, \text{Mel}_{refined}) \\
  &+ L_2(\text{Mel}_{gt}, \text{Mel}_{coarse}) + L_2(\text{Mel}_{gt}, \text{Mel}_{refined})
\end{aligned}
\end{equation}
To enhance audio reconstruction quality, we introduce a multi-scale Mel loss approach \cite{kumar2024high}, utilizing two distinct hop lengths and window sizes. This effectively mitigates information loss caused by dimensionality reduction and downsampling interpolation during the transformation from the decoder output to the Mel-spectrogram. By optimizing across multiple scales, the method preserves more detailed audio features, enhancing both reconstruction fidelity and training stability. For the pretrained LLM, the objective is to maximize the softmax probability of text outputs in audio understanding tasks. To ensure semantic alignment, the pretrained LLM maximizes the text softmax probability for audio understanding tasks, with its parameters kept fixed during training to preserve the alignment between the audio tokenizer and the textual LLM space. For the selection of LLM size, we observed during the training of audio understanding models that different LLM sizes have minimal impact on ASR performance metrics. Therefore, we chose a pretrained LLM with 1.5 billion parameters for continued pretraining. This LLM size also aligns well with the Audio Decoder, resulting in smaller gradient norm differences between the two components and improved training stability. We utilize an Exponential Moving Average (EMA) strategy to update the codebook and employ a Straight-Through Estimator (STE) for backpropagating gradients to the encoder. Additionally, we utilize a Vector Quantization (VQ) commitment loss to ensure the encoder outputs align closely with the codebook entries. The VQ commitment loss is defined as:
\begin{equation}
\text{Loss}_{commit} = | z - \text{quantize}(z) |_2^2,
\end{equation}
The total loss is a weighted combination of the multi-scale reconstruction loss, text-audio alignment loss (for the LLM), and VQ commitment loss:
\begin{equation}
\text{Loss}_{total} = \lambda_1 \cdot \text{Loss}^{'}_{reconstruct} + \lambda_2 \cdot \text{Loss}_{llm} +  \lambda_3 \cdot \text{Loss}_{commit}
\end{equation}
\textbf{RVQ Training Details.}
During the VQ training process, we introduce layerwise dropout, randomly discarding the VQ outputs of all layers starting from the second layer onward. This approach ensures that key semantic information is concentrated in the top layers of the codebook, achieving an effect similar to distilling the first-layer codebook in SpeechTokenizer \cite{zhang2024speechtokenizer}. To enhance codebook utilization, we implement a restart strategy alongside Gumbel sampling. Specifically, if a cluster within the codebook remains unused for a certain number of steps, it is randomly replaced with an input from the current batch. Meanwhile, we employ Gumbel-Softmax sampling during training to introduce stochasticity, ensuring effective exploration of the codebook space.  Additionally, to constrain the distribution of the codebook within a specific range, the L2 norm of the codebook is maintained during EMA updates. The final update formula is:
\begin{equation}
  c_{j, t} = c_{j, t-1} \times \alpha + \frac{1}{N} \sum_{x_{i} \in c_{j,t}} x_{i}
  \label{eq:formula1}
\end{equation}

\begin{equation}
  c_{j,t} = (1 - \beta) \times c_{j, t}^{\prime}
  \label{eq:formula2}
\end{equation}

where \( c_{j, t} \) denotes the codebook cluster at time step \( t \), \( c_{j, t-1} \) is its previous state, \( \alpha \) is the EMA decay factor, \( N \) is the number of inputs assigned to the cluster, and \( x_i \) represents the input samples in the current batch associated with \( c_{j, t} \). Here, \( c_{j, t}' \) represents the intermediate update of the cluster before normalization, and \( \beta \) is the constraint factor used to regulate the L2 norm of the codebook vector, ensuring its distribution remains within a bounded range. For the L2 norm constraint, \( c_{j, t}' \) is the intermediate updated cluster before normalization, and \( \beta \) is the constraint factor controlling the compression of the distribution of codebook vector. We observed that L2 norm constraints significantly reduce the mutual information of the codebook, concentrating the VQ distribution on a few top token IDs. This helps simplify downstream tasks, such as TTS, by improving the prediction accuracy of audio tokens. However, excessively strong L2 norm constraints can lead to codebook redundancy and reduce overall codebook utilization. Therefore, the $\beta$ parameter in the formula requires careful fine-tuning. To address this, we adopt a multi-stage progressive training strategy, gradually increasing the proportion of Whisper Encoder outputs replaced by VQ results, from 10\% to 100\%, while simultaneously adjusting the weight of the commit loss. We found that replacing the entire sample with VQ is more effective than randomly replacing individual tokens with VQ, i.e., instance-level replacement is better than token-level replacement.

\textbf{Training Data.}
In addition to traditional tasks such as Automatic Speech Recognition (ASR), Audio Query Answering (AQA), and Speech-to-Text Translation (S2TT), we incorporate a proportion of audio-text interleaved data into the training process. This strategy aims to enhance the VQ module's capability to model complex contextual scenarios. Specifically, the training dataset consists of 135k hours of ASR data, 11k hours of AQA data, 9k hours of S2TT translation data, and 52k hours of audio-text interleaved data. Further details about the dataset are provided in Section 3.3.

\textbf{Evaluation of Baichuan-Audio-Tokenizer.}
To evaluate the performance of the tokenizer in terms of pre-training and post-training losses, we trained a non-VQ version of the audio understanding model as a baseline using the same data and base model. For both the VQ and non-VQ models, the parameters of the LLM were kept frozen during training to ensure a fair comparison and to isolate the impact of the VQ mechanism on the overall performance. From \autoref{tab:vq8-metrics}, we can see that the 8-layer vq is closer to the baseline and has the least loss of semantic content. As shown in \autoref{tab:vq_metrics}, the ASR results of the 8-layer VQ model and the baseline across multiple datasets demonstrate that the trained 8-layer VQ model achieves competitive performance.

\begin{table}[!ht]
    \caption{\textbf{Comparison of ASR Performance across VQ Models with Different Layer Counts.} Librispeech* results are averages across dev-clean, dev-other, test-clean, and test-other. AiShell1* is the average of eval and dev results.}
    \label{tab:vq8-metrics}
    \centering
    \begin{tabular}{@{}lcccc@{}}
        \toprule
        \multicolumn{1}{c}{\textbf{Model}} &
          \begin{tabular}[c]{@{}c@{}}LibriSpeech*\\ WER (\%)\end{tabular} &
          \begin{tabular}[c]{@{}c@{}}AiShell1*\\ WER (\%)\end{tabular} &
          \begin{tabular}[c]{@{}c@{}}LibriSpeech*\\ Mel MAE\end{tabular} &
          \begin{tabular}[c]{@{}c@{}}AiShell1*\\ Mel MAE\end{tabular} \\ 
          \midrule
        Baseline                      & 3.8  & 2.0  & -      & -      \\ 
        8 Layers VQ           & 5.3  & 2.7  & 0.466  & 0.403  \\ 
        6 Layers VQ     & 5.5  & 3.0  & 0.485  & 0.423  \\ 
        4 Layers VQ     & 6.3  & 3.9  & 0.516  & 0.488  \\ 
        1 Layer VQ     & 26.2 & 19.2 & 0.677  & 0.553  \\ 
        \bottomrule
    \end{tabular}
\end{table}
\begin{table}[ht]
    \caption{\textbf{Comparison of ASR Performance between the Baseline and the 8-layer VQ Model.} The evaluation metric is WER(\%).}
    \centering
    \renewcommand{\arraystretch}{1.2}
    \begin{tabular}{lcc}
        \hline
        \textbf{Dataset} & \textbf{Baseline} & \textbf{VQ Model (8 Layers)} \\
        \hline
        Fleurs-ZH & 3.54 & 4.15 \\
        Fleurs-EN & 5.98 & 8.64 \\
        Med-ZH-inhouse & 4.39 & 6.03 \\
        Wenet testnet & 7.32 & 8.29 \\
        Wenet testmeeting & 8.54 & 9.03 \\
        Covost-2 en2zh BLEU & 33.94 & 30.4 \\
        Covost-2 zh2en BLEU & 21.7 & 19.7 \\
        Clotho AQA & 40.4 & 37.2 \\
        \hline
    \end{tabular}
    \label{tab:vq_metrics}
\end{table}
\subsection{Flow-matching based audio decoder}

\begin{wrapfigure}{r}{0.4\textwidth}
  \centering
  \includegraphics[width=3.6cm]{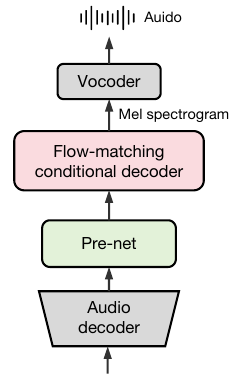}
  \caption{Flow-matching based audio decoder.}
  \label{fig:decoder}
\end{wrapfigure}

To improve the quality and fidelity of synthesized audio, the audio decoder module is enhanced using a flow-matching model \cite{lipman2022flow}, trained on 24 kHz audio to generate target Mel spectrograms. The flow-matching decoder includes a Pre-Net and a conditional decoder, as shown in \autoref{fig:decoder}. The Pre-Net maps intermediate representations to a prior for the vocoder, using an MLP and a 12-layer transformer. The MLP projects 1280-dimensional, 50 Hz features to 512 dimensions, refined by the transformer, and a final linear layer converts them into 80-dimensional Mel spectrograms. 
The flow-matching conditional decoder uses a U-Net structure trained with OT-CFM, inspired by Matcha-TTS \cite{mehta2024matcha} and CosyVoice \cite{du2024cosyvoice}. The U-Net includes one down-sampling block, one up-sampling block, and 12 intermediate blocks, each with a ResNet1D and transformer layer (256 dimensions). A linear layer projects features to 80-dimensional Mel spectrograms. As the model already encodes acoustic information (e.g., speaker timbre) via reconstruction loss, no additional speaker embeddings are used. The generated Mel spectrograms are converted into waveforms using the HiFi-GAN \cite{kong2020hifi,du2024cosyvoice} vocoder\footnote{https://www.modelscope.cn/models/iic/CosyVoice2-0.5B}.

\textbf{Training Details.}
The flow-matching model was trained on about 270k hours of audio, including Mandarin, English, various dialects, and multilingual data. Data quality was refined using ensemble ASR and MOS filtering. During training, the AudioEncoder, VQ layers, and AudioDecoder were fixed, while the flow-matching Pre-Net and decoder were trained with a prior loss added to the Pre-Net.

\textbf{Reconstruction result.}
We evaluated the performance of the trained flow-matching network on the LibriSpeech-dev set. UTMOS \cite{saeki2022utmos} was used as a proxy for subjective audio perception, with the score improving from 3.43 to 4.05, closely approaching the ground truth score of 4.08. Content quality was assessed using Whisper ASR \cite{radford2023robust}, where the Word Error Rate (WER) decreased from 2.84 to 2.78. 

\begin{table}[h]
    \caption{\textbf{Evaluation of reconstruction performance}. We evaluate the performance of subjective audio perception and content quality on librispeech-dev set.}
    \label{tab:fm-audio-data-summary}
\centering
\begin{tabular}{@{}lcc@{}}
\toprule
               & UTMOS\footnotemark[2] & ASR-WER \\ \midrule
groundtruth    & 4.08  & 2.26    \\
VQ             & 3.43  & 2.84    \\
+Flow-matching & 4.05  & 2.78    \\ \bottomrule
\end{tabular}
\end{table}

\subsection{Audio LLM}
\par The Baichuan-Audio extends a pre-trained LLM by incorporating the newly introduced Baichuan-Audio-Tokenizer, which includes audio embedding layers and an independent audio head. Specifically, the audio tokens from Baichuan-Audio-Tokenizer are first transformed into audio embeddings through audio embedding layers. The audio LLM alternately generates aligned text tokens and audio tokens, facilitated by a special token that enables modality switching between text and audio. The generated audio tokens are processed by an independent audio head, consisting of 3 layers of depth transformers and 8 classification heads. Finally, the audio embeddings are passed through an audio encoder, such as a flow-matching-based audio encoder and vocoder, to reconstruct audio waveforms. 
\footnotetext[2]{https://github.com/sarulab-speech/UTMOS22}

\textbf{Audio embedding.}
First, the 8 discrete audio token are summed through a corresponding number of embedding layers to obtain audio embeddings. Each embedding layer takes an input dimension that is one greater than the size of the corresponding codebook, due to the inclusion of an additional special token that signifies the end of audio token generation.

\textbf{Audio head.}
The generated audio tokens are processed using an independent audio head, which consists of 3 layers of depth transformers and 8 classification heads. The depth transformer has a depth of 8, predicting audio embeddings for 8 codebooks. Finally, the classification heads are used to obtain the logits for each codebook corresponding to the audio tokens.

Compared to pure text-based large models, speech language models often struggle to generate semantically coherent outputs. Research in \cite{wang2024speech} indicates that this issue primarily arises from the introduction of duration and paralinguistic information in speech. To address this issue, two types of interleaved data are employed during pre-training: Audio-Text Interleaved (INTLV) and Interleaved Text-to-Speech (ITTS), which contribute to enhancing audio understanding and generation capabilities. During inference, discrete audio tokens are fed into the LLM, and the model alternately generates aligned text tokens and audio tokens. Special tokens facilitate modality switching between text and audio. This forced alignment approach ensures that the model generates coherent and complete textual content before synthesizing the corresponding audio tokens, effectively guiding the generation of audio tokens and mitigating the issue of semantic degradation.

\begin{figure*}[htb]
  \centering
  \includegraphics[width=16cm]{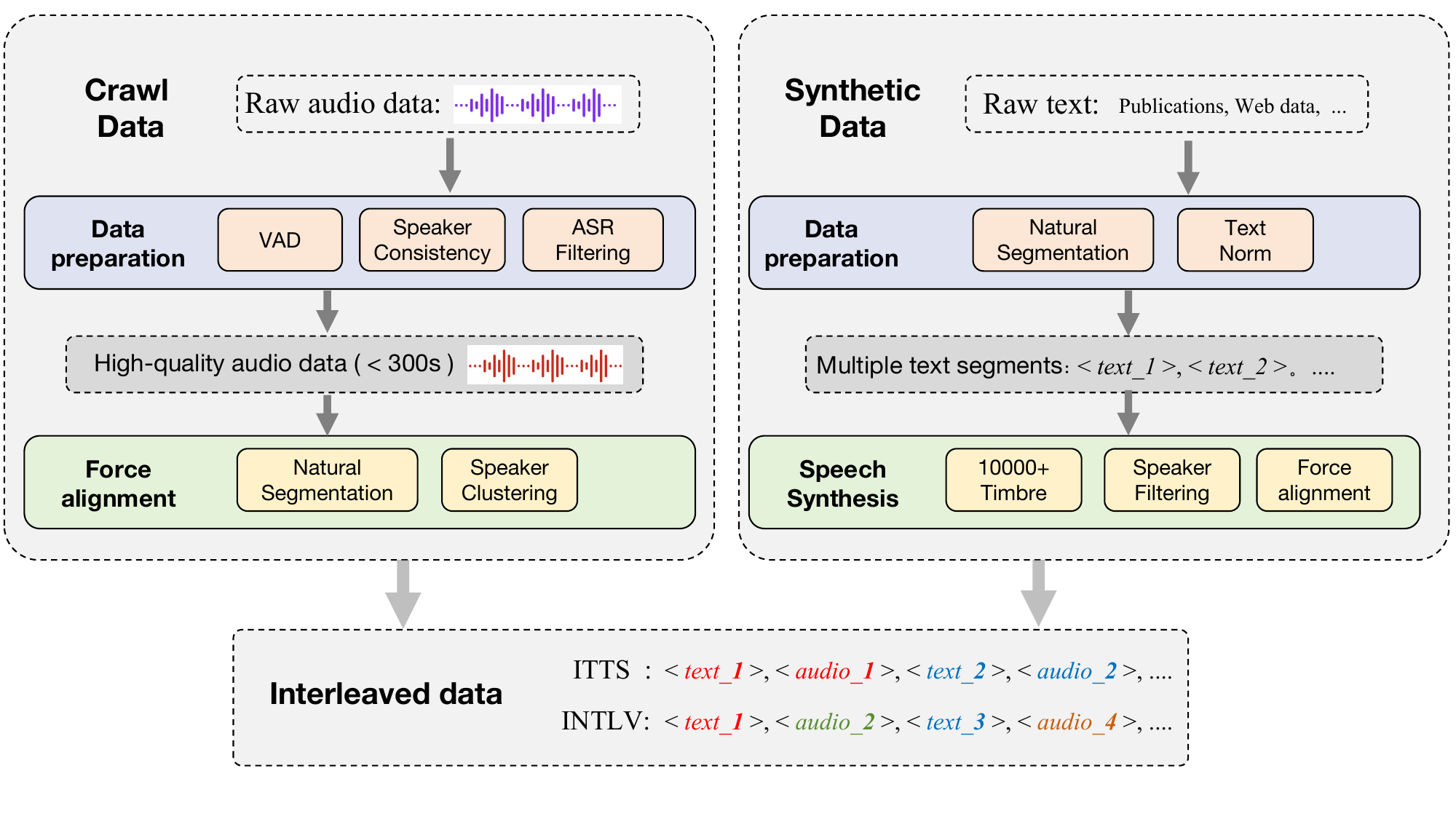}
  \caption{Pipeline of interleaved data collection.}
  \label{fig:data}
\end{figure*}
\subsubsection{Pre-training details}
\textbf{Pre-training data.}
\begin{table}[b]
    \caption{Detailed statistics of the training data of audio pretrain.}
    \label{tab:audio-data-summary}
    \resizebox{\textwidth}{!}{
    \centering
    \begin{tabular}{@{}lclcc@{}}
        \toprule
        Type & Task & Data Format & Hours (k) \\ 
        \midrule
        \multirow{4}{*}{Audio Understanding} 
          & Automatic Speech Recognition (ASR) & \texttt{<prompt, audio, transcript>} & 185 \\
          & Audio Query Answer (AQA) & \texttt{<prompt, audio, response>} & 21 \\
          & Speech-to-Text Translation (S2TT) & \texttt{<prompt, audio, translated\_text>} & 15 \\
          & Audio-Text Interleaved (INTLV) & \texttt{<audio\_1, text\_2, audio\_3, text\_4, ...>} & 393 \\ 
        \midrule
        \multirow{3}{*}{Audio Generation} 
          & Text-to-Speech (TTS) & \texttt{<text, audio>} & 51 \\
          & Interleaved Text-to-Speech (ITTS) & \texttt{<text\_1, audio\_1, text\_2, audio\_2, ...>} & 142 \\
          & Pure Audio & \texttt{<audio>} & 80 \\ 
        \midrule
        Total & - & - & 887 \\
        \bottomrule
    \end{tabular}}
\end{table}
The detailed statistics of the training data for audio pre-training are shown in \autoref{tab:audio-data-summary}. Audio data can be broadly categorized into two primary types: audio understanding data and audio generation data. Audio understanding data includes Automatic Speech Recognition (ASR), Audio Question Answering (AQA), Speech-to-Text Translation, and Audio-Text Interleave data. Audio generation data encompasses Text-to-Speech (TTS), Interleaved Text-to-Speech data, and pure audio data. Interleaved data consists of alternating text and audio modalities, segmented by punctuation marks to facilitate cross-modal knowledge transfer. The interleaved aligned generation data composed of fully aligned text and audio content, designed to enhance the model’s ability to generate audio tokens under text supervision. The audio-text paired data (e.g., ASR and TTS data) improve the performance on fundamental speech tasks. Pure audio data, on the other hand, enhances the capability to independently process audio modalities. The interleaved data collection process is shown in \autoref{fig:data}, which is divided into two types: crawl data and synthetic data. In total, we obtained 142k hours of ITTS data and 393k hours of INTLV data. The interleaved data is segmented using LLM, which performs natural segmentation based on punctuation or natural pauses in the text content. For the segmented text data of synthetic data, we also use a large language model for text normalization \cite{zhang2024chat}. During pre-training, we excluded the loss computation for the audio segments in the Audio-Text Interleaved data, a design choice that differs from GLM-4-Voice. Empirical observations under the current about 50B training audio data scale revealed that computing loss for the audio segments in INTLV data led to performance degradation. This decision is further justified by the inherent modality conflict between audio and text, as well as the absence of a requirement for text-to-audio continuation during inference. Consequently, we omitted the loss calculation for the audio segments in INTLV data. For ITTS data, loss was computed for both audio and text segments, except for the initial text segment, to enhance the model's capability in text-guided audio generation.

\textbf{Two stage training strategy.}
To address the potential disruption of the original textual knowledge in the LLM caused by the distinct characteristics of speech features compared to text features, we propose a two-stage training strategy to mitigate training conflicts between modalities. In the first stage, the parameters of the LLM remain fixed, allowing only the parameters of the audio embedding layer and the audio head to be updated. In the second stage, all parameters, except for those of the text embedding layer and the LM head, are made trainable.

\subsubsection{Supervised fine-tuning details}
The supervised fine-tuning stage aims to enhance the model's ability to follow complex instructions across a range of tasks. The audio SFT data are derived from a large collection of textual instructions. High-quality instructions are selected using a filtering strategy based on instruction type, diversity, and overall quality. Audio instructions are synthesized using a curated dataset of 10,000 distinct voice tones. Corresponding text responses are generated and segmented at natural conversational pauses before being converted into audio using the designated voice tones. These datasets cover multiple tasks and contain approximately 242k audio data pairs. 

To ensure the quality of the synthesized audio, Automatic Speech Recognition (ASR) is applied to the generated audio files. The ASR outputs are compared against the original text to validate quality. This process results in the creation of high-quality end-to-end conversational datasets. Synthesized audio files with errors are added to the Text-to-Speech (TTS) dataset, while cases with ASR errors are incorporated into the ASR training dataset. This iterative approach of incorporating challenging examples enhances both TTS and ASR performance.

Special attention is required to address cases where text-to-audio conversion makes the original textual response unsuitable as an audio reply. This issue arises due to differences in tone, speed, and expression between text and audio. Some textual content may fail to convey the intended meaning or introduce ambiguity when converted into audio. Consequently, careful review and adjustment of such cases are essential during the generation process. This ensures that the synthesized data accurately reflects real-world voice interaction scenarios, enhancing data reliability and improving the model's practical applicability.

\section{Experiment}

\subsection{General Intelligence Evaluation}
A major challenge faced by speech-based dialogue models is that, compared to pure text-based dialogue models, their performance tends to degrade. To evaluate the "intelligence" of speech models, we use text-to-text modeling capabilities as a baseline and assess the performance of pre-trained speech-to-text models. The evaluation dataset consists of two types: story continuation ability and commonsense reasoning ability.

\textbf{Sentence continuation.}
For the continuation ability evaluation, we use the sStoryCloze dataset \cite{hassid2024textually}. Additionally, we introduce the zh-sStoryCloze dataset, which is created by translating the English version of sStoryCloze into its Chinese counterpart via a translation engine and replacing English names with Chinese ones to better suit the Chinese context. Each sample in both evaluation sets consists of five sentences, divided into positive and negative samples. The last sentence differs between the two, with the last sentence of the positive sample being the correct continuation. A prediction is considered correct if the perplexity of the last sentence in the positive sample is lower than that of the negative sample.

\textbf{Commonsense reasoning.}
For the commonsense reasoning ability evaluation, the goal is to assess whether the model possesses domain-specific knowledge. Drawing inspiration from the design of sStoryCloze, we use the GPT-4o API to rewrite and filter the CMMLU dataset \cite{li2023cmmlu}, ultimately creating the sCMMLU dataset with 4,743 commonsense questions. For each multiple-choice question in the original CMMLU, we rewrite it into four statements with the same first half and different second halves according to the answer options. A prediction is considered correct if the perplexity of the correct option's statement is lower than that of the other options.

\begin{table}[htb]
    \caption{\textbf{Performance Comparison on Various Evaluation Tasks.} $\ast$: Evaluations were performed using the instruct model as no base model was provided.}
    \label{tab:iq-performance}
    \centering
    \begin{tabular}{@{}lccccc@{}}
        \toprule
        \multirow{2}{*}{\textbf{Model}}   & \multirow{2}{*}{\textbf{Modality}} & \multirow{2}{*}{\textbf{Params}} & \multicolumn{3}{c}{\textbf{Evaluation Datasets}} \\ \cmidrule(l){4-6}
                         &                   &                 & \textbf{sStoryCloze} & \textbf{zh-sStoryCloze} & \textbf{sCMMLU} \\ \midrule
        TWIST            & S $\to$ T         & 7B               & 53.3                  & -                       & -              \\
        Moshi            & S $\to$ T         & 7B               & 60.8                  & -                       & -              \\
        GLM-4-Voice      & S $\to$ T         & 9B               & 76.3                  & 70.3$^{\ast}$                       & 64.3$^{\ast}$              \\
        \midrule
        Baseline         & T $\to$ T         & 7B               & 83.0                  & 76.1                    & 70.3           \\
        Our (single stage)  & S $\to$ T         & 7B               & 77.5                  & 70.1                    & 67.0           \\
        Our (two stage)     & S $\to$ T         & 7B               & 79.6                  & 72.4                    & 69.3           \\ \bottomrule
    \end{tabular}
\end{table}

\textbf{Results.}
The general intelligence evaluation results in two aspects are shown in~\autoref{tab:iq-performance}. It can be observed that Baichuan-Audio consistently outperforms previous models in the sentence continuation evaluation task. Given the current architecture, the intelligence of the speech multimodal model is inherently dependent on the reasoning capability of the pure text LLM. Consequently, the accuracy of the T$\to$T mode serves as the upper bound for end-to-end speech models, while the accuracy of the S$\to$T mode is inherently lower than that of the T$\to$T mode. Our goal is to bridge this gap by improving the accuracy of the S$\to$T mode to approach that of the T$\to$T mode, thereby enhancing the intelligence of end-to-end speech models. Moreover, we observe that a two-stage training strategy effectively mitigates the degradation in model intelligence compared to single-stage training.

\subsection{Performance in ASR/TTS Tasks}
For ASR evaluation in the general scene, we report results on the Fleurs~\citep{fleurs2022arxiv} Chinese (\textit{zh}) and English (\textit{en}) test sets, as well as the WenetSpeech~\citep{zhang2022wenetspeech10000hoursmultidomain} \textit{test\_net} dataset. To assess performance in more challenging ASR scenarios,  we include results from the WenetSpeech~\citep{zhang2022wenetspeech10000hoursmultidomain} \textit{test\_meeting} dataset and the KeSpeech~\citep{tang2021kespeech} test set, which evaluate the model's ASR capabilities in `Meeting' and `Chinese dialect' contexts. Baichuan-Audio exhibits a strong audio transcription capacity in~\autoref{tab:audio-general-asr}. On the Fleurs dataset (test-zh), it achieves a WER of 3.2\%, significantly lower than Whisper-large-v3 (12.4\%) and Qwen2-Audio-Base (4.3\%). For the WenetSpeech dataset, Baichuan-Audio-Base achieves a WER of 7.2\% on test\_net and 8.5\% on test\_meeting. On the KeSpeech dataset, Baichuan-Audio-Base excels across multiple Chinese dialects. In addition to ASR, Baichuan-Audio excels in both S2TT and TTS tasks. For S2TT task, which aims to translate the audio signal in the source to the target language. We evaluate the model's S2TT performance between Chinese and English using the zh2en and en2zh subsets of the Covost2~\citep{wang2020covost2massivelymultilingual} dataset, with BLEU~\citep{papineni-etal-2002-bleu} scores as the evaluation metric. The evaluation results of S2TT and TTS tasks are summarized in~\autoref{tab:bleu-tts}. 
\begin{table}[ht]  
\centering  
\caption{\textbf{Major results on} \textbf{Fleurs}, \textbf{WenetSpeech}, \textbf{and} \textbf{KeSpeech}. The test sets are evaluated with WER. The rest unlabeled results are reproduced by ourselves, and any performance divergence may be attributed to differences in decoding parameters.}  
\begin{tabular}{lccc}  
\toprule  
\multicolumn{1}{c}{\multirow{2}{*}{\textbf{Scene}}} & \multirow{2}{*}{\textbf{Dataset}} & \multirow{2}{*}{\textbf{Model}} & \multirow{2}{*}{\begin{tabular}[c]{@{}c@{}}\textbf{Results} \\ WER (\%) $\downarrow$\end{tabular}} \\
\multicolumn{1}{c}{} & & & \\   
\midrule  
\multirow{7}{*}{General} & \multirow{3}{*}{\begin{tabular}[c]{@{}c@{}}Fleurs \\ \textit{test-zh} \end{tabular}} & Whisper-large-v3 (1.55B) & 12.4  \\
& & Qwen2-Audio-Base (7B) & 4.3  \\ 
& & Baichuan-Audio-Base (7B)& \textbf{3.2 } \\
\cmidrule(l){2-4}   
& \multirow{4}{*}{\begin{tabular}[c]{@{}c@{}}WenetSpeech \\ \textit{test\_net}\end{tabular}} & Whisper-large-v3 (1.55B) & 17.5  \\
& & Qwen2-Audio-Base (7B) & 7.3  \\
& & Baichuan-Audio-Base (7B)& \textbf{7.2 } \\ 
\midrule  
\multirow{4}{*}{Meeting} & \multirow{4}{*}{\begin{tabular}[c]{@{}c@{}}WenetSpeech \\ \textit{test\_meeting}\end{tabular}} & Whisper-large-v3 (1.55B) & 30.8 \\
& & Qwen2-Audio-Base (7B) & \textbf{7.7} \\
& & Baichuan-Audio-Base (7B)& 8.5 \\
\midrule  
\multirow{6}{*}{\begin{tabular}[c]{@{}l@{}}Chinese \\ dialect\end{tabular}} & \multirow{6}{*}{\begin{tabular}[c]{@{}c@{}}KeSpeech\\ \textit{mandarin} | \textit{beijing} | \textit{southwest}\\ \textit{lan-yin} | \textit{zhongyuan} | \textit{northeast}\\ \textit{jiang-huai} | \textit{ji-lu} | \textit{jiao-liao}\end{tabular}} & Whisper-large-v3 (1.55B) & \begin{tabular}[c]{@{}c@{}}18.7 | 44.8 | 52.9\\ 54.8 | 50.1 | 22.9\\ 54.7 | 47.0 | 50.4\end{tabular} \\
\cmidrule(l){3-4}  
& & Qwen2-Audio-Base (7B) & \begin{tabular}[c]{@{}c@{}}3.0 | 7.5 | \textbf{6.2}\\ \textbf{6.7} | \textbf{5.0} | 5.5\\ \textbf{9.1} | \textbf{6.6} | \textbf{7.0}\end{tabular} \\

\cmidrule(l){3-4}  
& & Baichuan-Audio-Base (7B)& \begin{tabular}[c]{@{}c@{}}\textbf{2.7} | \textbf{6.9} | 6.8\\ 7.2 | 5.3 | \textbf{5.0}\\ 9.9 | 6.7 | 7.4\end{tabular} \\ 
\bottomrule   
\end{tabular}  
\label{tab:audio-general-asr}  
\end{table}
\begin{table}[htbp]
\caption{\textbf{Evaluation the capabilities of the automatic speech translation and text-to-speech on base model}.}
\centering
\begin{tabular}{cccc}
\toprule
\multirow{2}{*}{\textbf{Model}} & \multicolumn{2}{c}{\textbf{Translation (BLEU)}} & \textbf{TTS (ASR)} \\
\cmidrule(lr){2-3} \cmidrule(lr){4-4}
 & Covost-2 zh-CN2en & Covost-2 en2zh-CN & MED-TTS \\
\midrule
\textbf{Qwen2-Audio-Base (7B)} & 22.17 & 43.58 & - \\
\textbf{Baichuan-Audio-Base (7B)} & 24.37 & 45.96 & 2.71 \\
\bottomrule
\end{tabular}
\label{tab:bleu-tts}
\end{table}

\subsection{Performance in Audio Understanding Tasks}

\textbf{Baselines.} We compare Baichuan-Audio with the following baselines: proprietary model (GPT-4o-Audio~\cite{HelloGPT4o}), open-source voice model (GLM-4-Voice~\cite{zeng2024glm}), and open-source models for omni-modal (VITA-1.5~\cite{fu2025vita}, MiniCPM-o 2.6~\cite{yao2024minicpm}).

\textbf{Evaluation Benchmarks.} To assess the audio understanding capabilities of Baichuan-Audio, we have built and open-sourced an OpenAudioBench and use GPT-4o~\cite{HelloGPT4o} to evaluate the results, including Reasoning QA(self-constructed), Spoken Llama Questions~\cite{nachmani2023spoken}, Web Questions~\cite{berant2013semantic}, TriviaQA~\cite{joshi2017triviaqa}, and AlpacaEval~\cite{li2023alpacaeval}. For AlpacaEval, we select two subsets \texttt{helpful base} and \texttt{vicuna} from the original AlpacaEval dataset and remove questions related to math and code. This process follows Llama-Omni~\cite{fang2024llama}, with the aim of obtaining questions more suitable for speech scenarios, and the final AlpacaEval benchmark in our report comprises 199 questions in total. Considering the substantial size of the Web Questions and TriviaQA datasets, a full evaluation is impractical. Therefore, we randomly sample 1,000 questions from each original dataset. The instructions for these three benchmarks were synthesized using our TTS model.

For Reasoning QA, we use GPT-4o to evaluate the score of the answers based on the given reference answers, and then calculate the accuracy rate. For Llama Questions, Web Questions, and TriviaQA, we provide reference answers and use GPT-4o to assess the correctness of the model's responses. The final score is the percentage of answers judged as correct.

For all audio benchmarks, we consider two different settings: 1) speech-to-speech generation in a non cascaded manner (denoted as S$\to$S), where the input is audio and the output is interleaved text and audio. The output text is then merged and used for evaluation. 2) speech-to-text generation (denoted as S$\to$T, where the input is audio and the output is text, which is used for evaluation.

\textbf{Results.} As shown in \autoref{tab:audio bench}, our model performs excellently on audio understanding benchmarks, outperforming the latest open-source models. In the S$\to$T setting, Baichuan-Audio significantly outperforms models of the same size in AlpacaEval, achieving score of 77.4. In the S$\to$S setting, Baichuan-Audio surpasses GLM-4-Voice across the board, particularly leading by 11.4 and 20.7 in Reasoning QA and AlpacaEval.

\begin{table}[!ht]
    \caption{\textbf{Results on audio understanding benchmarks.} $\nabla$: The modalities parameter is set to ["text", "audio"], evaluation based on the output text. $\diamondsuit$: Supports only text-audio interleaved output. $\square$: Cascade output method, evaluation based on the output text.}
    \label{tab:audio bench}
    \resizebox{\textwidth}{!}{
    \centering
    \begin{tabular}{@{}ccccccccccc@{}}
        \toprule
        \multicolumn{1}{c|}{}    & \multicolumn{10}{c}{\textbf{Audio Comprehensive Capacity}} \\
        \cmidrule{2-11}
        \multicolumn{1}{c|}{\multirow{2}{*}{\textbf{Model}}} & \multicolumn{2}{c|}{Reasoning QA} & \multicolumn{2}{c|}{Llama Questions} & \multicolumn{2}{c|}{Web Questions} & \multicolumn{2}{c|}{TriviaQA} & \multicolumn{2}{c}{AlpacaEval} \\
        \cmidrule(l){2-11}
         \multicolumn{1}{c|}{} & \textit{s $\to$ t} & \multicolumn{1}{c|}{\textit{s $\to$ s}} & \textit{s $\to$ t} & \multicolumn{1}{c|}{\textit{s $\to$ s}} & \textit{s $\to$ t} & \multicolumn{1}{c|}{\textit{s $\to$ s}} & \textit{s $\to$ t} & \multicolumn{1}{c|}{\textit{s $\to$ s}} & \textit{s $\to$ t} & \textit{s $\to$ s} \\
          \midrule
         \multicolumn{1}{c|}{GPT-4o-Audio$^\nabla$} & \textbf{55.6} & \multicolumn{1}{c|}{-} & \textbf{88.4} & \multicolumn{1}{c|}{-} & \textbf{81.0} & \multicolumn{1}{c|}{-} & \textbf{90.6} & \multicolumn{1}{c|}{-} & \textbf{80.1} & - \\
         \multicolumn{1}{c|}{GLM-4-Voice (9B)$^\diamondsuit$} & - & \multicolumn{1}{c|}{26.5} & - & \multicolumn{1}{c|}{71.0} & - & \multicolumn{1}{c|}{51.5} & - & \multicolumn{1}{c|}{46.6} & - & 48.9 \\
         \multicolumn{1}{c|}{VITA-1.5 (7B)$^\square$} & 41.0 & \multicolumn{1}{c|}{-} & 74.2 & \multicolumn{1}{c|}{-} & 57.3 & \multicolumn{1}{c|}{-} & 46.8 & \multicolumn{1}{c|}{-} & 68.2 & - \\
        \multicolumn{1}{c|}{MiniCPM-o 2.6 (7B)$^\square$} & 38.6 & \multicolumn{1}{c|}{-} & 77.8 & \multicolumn{1}{c|}{-} & 68.6 & \multicolumn{1}{c|}{-} & 61.9 & \multicolumn{1}{c|}{-} & 51.8 & - \\
         \multicolumn{1}{c|}{\textbf{Baichuan-Audio (7B)}} & 41.9 & \multicolumn{1}{c|}{\textbf{37.9}} & 78.4 & \multicolumn{1}{c|}{\textbf{74.5}} & 64.5 & \multicolumn{1}{c|}{\textbf{60.3}} & 61.7 & \multicolumn{1}{c|}{\textbf{54.2}} & 77.4 & \textbf{69.6} \\
        \bottomrule
    \end{tabular}}
\end{table}

\section{Conclusion}

In this paper, we introduce Baichuan-Audio, an end-to-end large language model designed for audio that integrates both speech comprehension and generation. The model employs a multi-codebook discretization of speech signals at 12.5 Hz via a pre-trained ASR model, which preserves both semantic and acoustic information in speech tokens. Additionally, an independent audio head is specifically designed to process these tokens efficiently. To balance audio modeling and language capability preservation, a two-stage pre-training strategy with interleaved data is adopted. The proposed framework supports speech interaction through text-guided aligned speech generation, thereby further retaining the model's foundational cognitive abilities. With open-sourced training data and models, Baichuan-Audio makes a significant contribution to the advancement of real-time speech interaction systems.

\bibliography{references}
\bibliographystyle{plain}

\clearpage
\appendix
\section{Appendix}

\subsection{Open-Source Data for training}

\begin{table}[ht]
\centering
\caption{Open-Source Dataset Summary}
\label{table:datasets}
\begin{tabular}{l l l}  
\toprule
Dataset & Type & Sizes  \\
\midrule
AISHELL 1 & Chinese & 500 hours  \\ 
AMI-IHM/SDM & English & 100 hours \\ 
AudioCaps & English & 46,000 audios   \\ 
AudioSet & Multilingual & 5,800 hours   \\ 
Clotho & English & 4981 audios   \\ 
CMU-MOSEI & - & 65 hours  \\ 
Common Voice & Multilingual & 31,000   \\ 
covost2 & Multilingual & 2900   \\ 
Emilia & Multilingual & 101k   \\ 
earnings22 & English & 119 hours (57k clips)   \\ 
Fluent Speech Commands & English & 10 hours  \\ 
fma & Music & -   \\ 
fsd50k & - & 51k clips (108h)   \\ 
GigaSpeech & English & 10,000 hours   \\ 
gigaspeech2 & Southeast Asian & 30,000 hours   \\ 
Google FLEURS & Multilingual & 10 hours   \\ 
kespeech & Chinese Dialects & 1,500+ hours   \\ 
LibriSpeech & English & 1,000 hours   \\ 
LibriTTS & English & 585 hours   \\ 
LJSpeech & English & 24 hours   \\ 
MAGICDATA & Chinese & 700+ hours   \\ 
Multilingual LibriSpeech & Multilingual & -   \\ 
Multilingual TEDx & Multilingual & -   \\ 
NMSQA\_audio & English & -   \\ 
Parler TTS & Multilingual & 54k hours  \\ 
peoples\_speech & Multilingual & 30,000+ hours   \\ 
SPGISpeech & English & 5,000+ hours   \\ 
TED-LIUM & English & 452 hours   \\ 
vggsound & - & 550+ hours   \\ 
wenetspeech4tts & Chinese & 12.8k+ hours   \\ 
WenetSpeech & Chinese & 10,000 hours   \\ 
zhvoice & Chinese & 960 hours   \\ 
\bottomrule
\end{tabular}

\end{table}

\end{document}